\documentclass[10pt,twocolumn,letterpaper]{article}

\usepackage{iccv}
\usepackage{times}
\usepackage{epsfig}
\usepackage{graphicx}
\usepackage{amsmath}
\usepackage{amssymb}

\usepackage{subfigure}
\usepackage{multirow}
\usepackage{booktabs}
\usepackage{xcolor}
\usepackage[normalem]{ulem}
\usepackage{siunitx}

\newlength\savewidth\newcommand\shline{\noalign{\global\savewidth\arrayrulewidth
		\global\arrayrulewidth 1pt}\hline\noalign{\global\arrayrulewidth\savewidth}}
\usepackage[pagebackref=true,breaklinks=true,letterpaper=true,colorlinks,bookmarks=false]{hyperref}

\iccvfinalcopy 

\def\iccvPaperID{11} 

\makeatletter\renewcommand\paragraph{\@startsection{paragraph}{4}{\z@}
	{.5em \@plus1ex \@minus.2ex}{-.5em}{\normalfont\normalsize\bfseries}}\makeatother

\ificcvfinal\fi
\begin{document}

\title{Learning From Synthetic Photorealistic Raindrop for\\Single Image Raindrop Removal}

\author{
    Zhixiang Hao$^{1}$\quad
    Shaodi You$^{3}$\quad
    Yu Li$^{4}$\quad
    Kunming Li$^{5}$\quad
    Feng Lu$^{1,2,}$\thanks{Corresponding Author: Feng Lu}
    \\
    $^{1}$State Key Laboratory of VR Technology and Systems, Beihang University, Beijing, China\\
    $^{2}$Peng Cheng Laboratory, Shenzhen, China,
    $^{3}$Data61-CSIRO,
    $^{4}$Tencent,
    $^{5}$Australian National University
    \\
    {\tt\small \{haozx, lufeng\}@buaa.edu.cn,}
    {\tt\small youshaodi@gmail.com,}
    {\tt\small yul@illinois.edu,}
    {\tt\small u5580030@alumni.anu.edu.au}
}

\maketitle

\begin{abstract}
	Raindrops adhered to camera lens or windshield are inevitable in rainy scenes and can become an issue for many computer vision systems such as autonomous driving. Because raindrop appearance is affected by too many parameters, it is unlikely to find an effective model based solution. Learning based methods are also problematic, because traditional learning method cannot properly model the complex appearance. Whereas deep learning method lacks sufficiently large and realistic training data. To solve it, in our work, we propose the first photo-realistic dataset of synthetic adherent raindrops with pixel-level mask for training. The rendering is physics based with consideration of the water dynamic, geometric and photometry. The dataset contains various types of rainy scenes and particularly the rainy driving scenes. Based on the modeling of raindrop imagery, we introduce a detection network which has the awareness of the raindrop refraction as well as its blurring. Based on that, we propose the removal network that can well recover the image structure. Rigorous experiments demonstrate the state-of-the-art performance of our proposed framework.
\end{abstract}
\let\thefootnote\relax\footnote{This work was supported by the National Natural Science Foundation of China (NSFC) under Grant 61972012 and Grant 61732016.}

\section{Introduction}


\begin{figure}
	\centering
	\subfigure[Real-world raindrop image]{
		\hspace{-0.8mm}\includegraphics[width=0.48\linewidth]{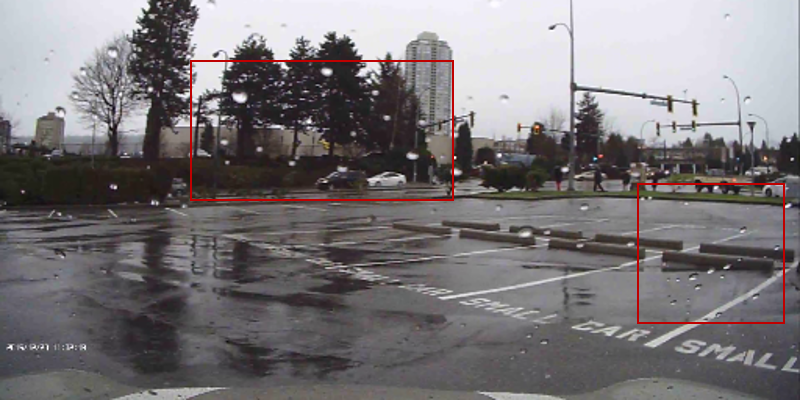}}\hfill
	\subfigure[Ours]{
		\includegraphics[width=0.48\linewidth]{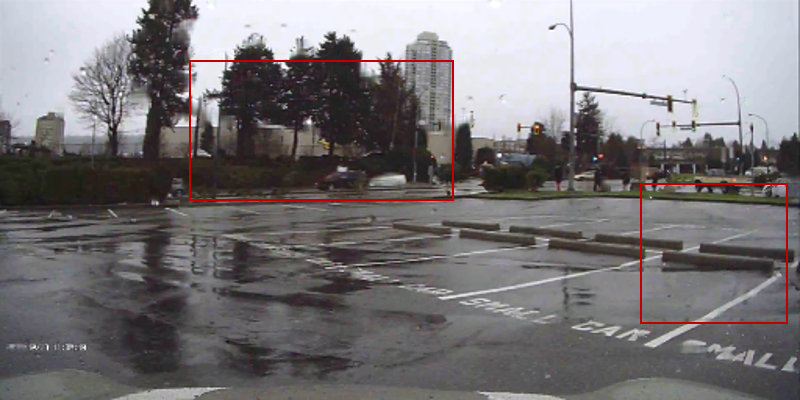}}\vfill\vspace{-2.0mm}
	\subfigure[Qian \cite{qian2018attentive}]{
		\hspace{-0.8mm}\includegraphics[width=0.48\linewidth]{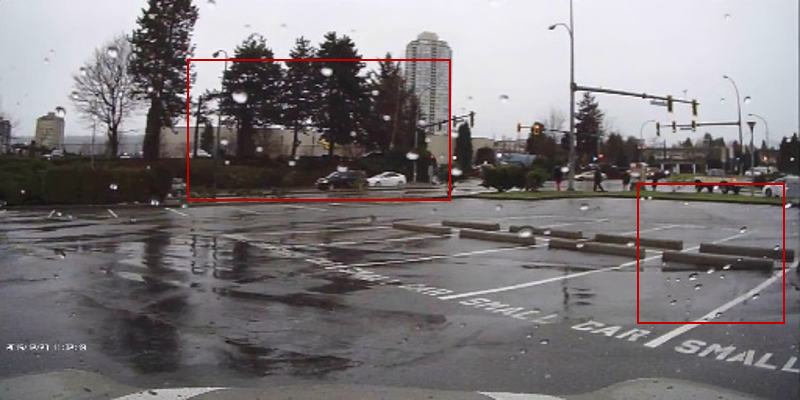}}\hfill
	\subfigure[Pix2Pix \cite{isola2017image}]{
		\includegraphics[width=0.48\linewidth]{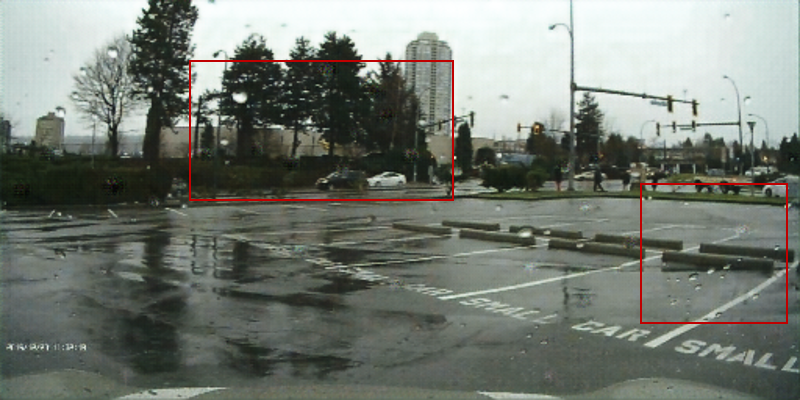}}
	\caption{Visual comparison of raindrop removal in real rainy scenes. Our method removes most of raindrops although the raindrops have large variety.}
	\label{fig:teaser}
\end{figure}

Most computer vision studies assume that the input image is of good visibility and clean content. However, rainy weather causes several different types of degradation to the image captured. It is common that the raindrops hit and flow on a camera lens or a windscreen of the vehicle. These adherent raindrops can obstruct, deform, and/or blur part of the area in the imagery of the background scenes, and then significantly degrade the performances of many vision algorithms \eg feature detection~\cite{ren2015faster, he2017mask, redmon2017yolo9000}, tracking~\cite{wu2013online, danelljan2017eco, valmadre2017end}, stereo correspondence~\cite{scharstein2002taxonomy, seitz2006comparison, godard2017unsupervised}, \etc. A method to automatically remove the raindrop and recover the clear scene is, therefore, desired.

Unlike the rain streaks \cite{yang2017deep,li2016rain},  that mostly are thin and vertical stripes, adherent raindrops have more varieties in shape, position, and size, as can be seem in Fig.~\ref{fig:teaser}a. 
You~\etal \cite{you2016adherent,you2014raindrop,you2013adherent}, Roser~\etal \cite{roser2009video}, Eigen~\etal \cite{eigen2013restoring} and Qian~\etal \cite{qian2018attentive} are a few example approaches focusing on detecting or removing the adherent raindrops. 
However, the method in \cite{you2016adherent} requires the rich temporal information, whereas the required video sequence cannot be applied to single image. Roser~\etal's method \cite{roser2009video} can detect raindrop from a single image, but the model is over simplified and far from the real cases. Rather than model based methods, Eigen~\etal \cite{eigen2013restoring} first adopt deep neural network, but the network only contains three layers and cannot properly learn the appearance of real raindrops. Qian~\etal \cite{qian2018attentive} integrate attention mechanism into GAN based CNNs, but their method is only tested on a small dataset. Although their dataset uses real raindrop, but the scene is in sunny day, which is not realistic. And therefore their method cannot fully handle real rainy scenes (Fig.~\ref{fig:teaser}c).

In this paper, we propose a physics driven as well as data driven method to detect and remove the adherent raindrops jointly.
We utilize the realistic adherent raindrops imagery model proposed by Roser \etal\cite{roser2010realistic} and You \etal\cite{you2016adherent}.
Based on understanding the physics, we design a novel deep learning multi-tasks network which does both end to end detection and removal adherent raindrops from a single image. Unlike existing networks, the proposed network directly reflects the appearance of raindrop such that it is partially blended into the image and is a reflection of the background image.
In brief, we separate the difficult task of restoring image into three sub-problems: (\emph{i}) detect raindrop locations and shapes via a deeply supervised sub-network, and then (\emph{ii}) restore adherent raindrop regions through deep learning network, subsequently (\emph{iii}) a small CNN network is employed to smooth the blended image. 

To enable proper training of the network, a new dataset is introduced which consisting photo-realistic rendering of the rainy scenes and clear scenes. 
The dataset uses Cityscapes dataset \cite{cordts2016cityscapes} as background image, which contains representative outdoor scenes. The dataset contains about $30K$ images. Each image has 50 to 70 raindrops with size varying from 0.8 to 1.5 centimeters, and the blurring level varying from 7 to 20 pixel.

This paper makes the following contributions:
\begin{itemize}
	\vspace{-1mm}\item We propose a physics aware end-to-end neural network for joint raindrop detection and removal. The architecture is designed in cope with the physics of raindrop imagery.
	
	\vspace{-2mm}\item We develop a practical dataset of realistically rendered adherent raindrop images, which contains the pixel-level raindrop binary masks.
	
	\vspace{-2mm}\item The proposed method significantly out performances existing methods on all existing dataset and real-world rainy images.
\end{itemize}

\section{Related Work}

Removing raindrops from a single image is an ill-posed problem and would be beneficial to outdoor computer vision systems which work in bad weather, particularly surveillance systems and intelligent vehicle systems.  Although there are many papers focus on removing haze \cite{he2011single,cai2016dehazenet} or rain streaks \cite{luo2015removing, li2016rain, zhang2018density}, the researches on raindrop removal from a single image are relatively insufficient. 

\subsection{Adherent Raindrop Modeling}

Halimeh~\etal \cite{halimeh2009raindrop} introduce a raindrop modeling method based on ray-tracking. They propose an algorithm which models the geometric shape of a raindrop by utilizing its photometric properties. Roser~\etal \cite{roser2010realistic} mainly focus on modeling the raindrop geometric shape. They leverage the B\'ezier curves to represent a raindrop surface in low dimensions which is physically interpretable. Von Bernuth~\etal \cite{von2018rendering} propose a novel method to render these raindrops using Continuous Nearest Neighbor search leveraging the benefits of R-trees. They use the synthetic raindrops for robustness verification of camera-based object recognition.

Recently, You~\etal \cite{you2016waterdrop} model raindrops by considering both liquid dynamics and optics. They reconstruct the 3D geometry of a raindrop by minimizing surface energy constraints and total reflection constraint. The accurate raindrop model proposed by You~\etal can be used in applications such as depth estimation and image refocusing. Later, You~\etal \cite{you2016adherent} model adherent raindrops by taking consideration of physical properties such as gravity, water-water surface tensor and water-adhering-surface tensor. 

\subsection{Raindrop Removal}

Most existing methods for detecting or removing raindrops are stereo or video based and therefore not applicable to a single image. Roser and Geiger \cite{roser2009video} propose a method which detects raindrop in a single image based on a photometric raindrop model. The raindrop detection can improve image registration accuracy, then removing raindrops by fusing multiple views into one frame. You~\etal \cite{you2016adherent} combine video completion technique with temporal intensity derivative to remove raindrops in video after detecting the locations of raindrops.

Due to the lack of temporal information, raindrop removal from a single image is more challenging. Eigen~\etal's work \cite{eigen2013restoring} is the first one to remove raindrops from a single image. They propose a 3-layer CNN network trained on rainy/clear pairs, the network can remove relatively sparse and small raindrops as well as dirt. However, the method suffers from blurred outputs and cannot remove dense raindrops. Recently, Qian~\etal \cite{qian2018attentive} propose a method based on GAN \cite{goodfellow2014generative}. They create an aligned dataset by using a piece of glass sprayed with water to get images containing raindrops. With this dataset, they propose a GAN based network which integrates attention mechanism both in generator and discriminator. The method can produce sharp and clear image on their test set.

There are also some general Image-to-Image translation methods such as Pix2Pix \cite{isola2017image} can tackle this problem, but they are not specifically designed for raindrop removal from a single image.

\section{Raindrop Imagery Model and Photorealistic Dataset}

As preliminary, we briefly introduce the raindrop imagery model developed by Roser \etal \cite{roser2010realistic} and extended by You \etal \cite{you2016adherent} and the implementation detail on our photorealistic dataset generated from such model. It will later drive us to design the network structure in Sec.~\ref{sec:mehtod}. Also, we introduce the detail of the new photo realistic dataset.

\paragraph{Motivation:} Data driven methods, particularly deep neural networks, need a large training data with ground truth. 
In particular, we need images with raindrops and the corresponding clear images to perform supervised learning in the context of raindrop removal from a single image. 
However, it is difficult and expensive to get strictly aligned rainy/clear image pairs of the exact same scene. Qian~\etal \cite{qian2018attentive} create a dataset contains $1119$ pairs in total. The dataset is the only one for adherent raindrops, but it is relatively small and lacks the pixel-level masks of raindrops. In order to train our network, we create the first photo-realistic adherent raindrop dataset with pixel-level mask in autonomous driving settings based on Cityscapes dataset \cite{cordts2016cityscapes}. Inspired by \cite{halimeh2009raindrop} and \cite{roser2010realistic}, we synthesize adherent raindrop appearance on a clear background image by tracking the ray from camera to environment through the raindrops.

\begin{figure}[t]
	\begin{center}
		\includegraphics[width=1.0\linewidth]{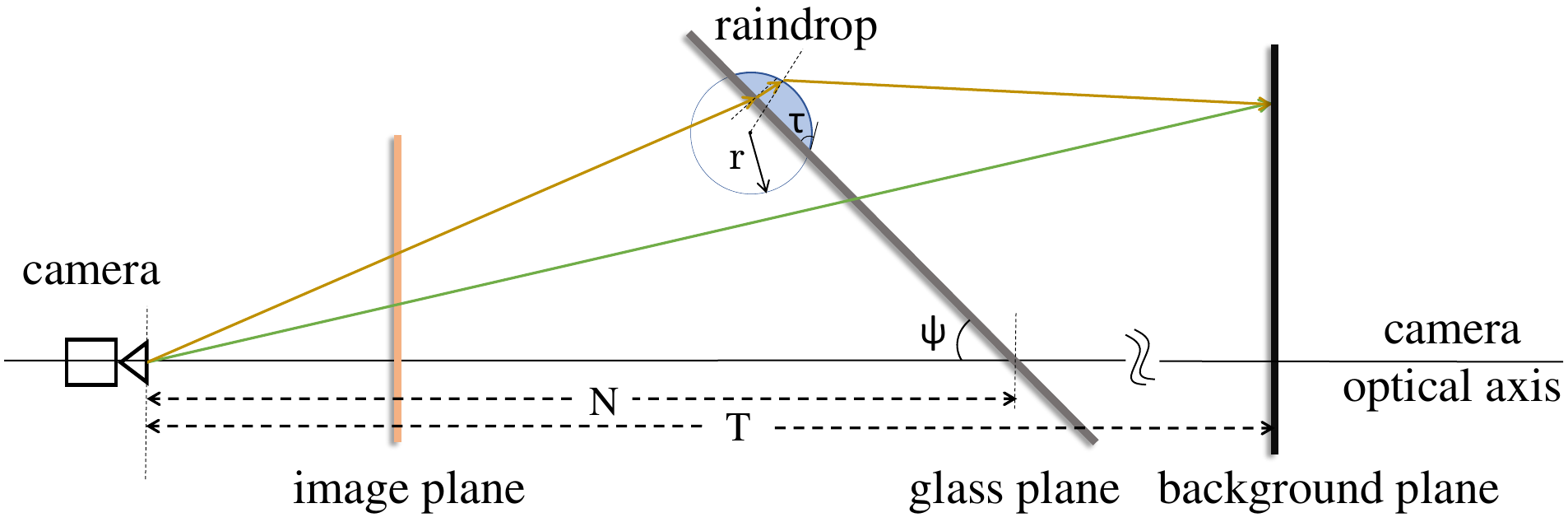}
	\end{center}
	\caption{Refraction model. The light ray colored in green does not go through any raindrops. The light ray colored in yellow goes through a raindrop and is refracted twice.}
	\label{fig:refraction}
\end{figure}

\paragraph{Dataset Generation: Geometric Rendering and Ray-tracing.} As shown in Fig.~\ref{fig:refraction}, in order to get the synthetic adherent raindrop images, we set a scene with a camera at the origin, a glass plane at $N$ centimeters ahead the camera, and a background plane at $T$ centimeters ahead the camera. The angle between the glass plane and the ground is $ \psi $. On the glass plane, we randomly sprinkle raindrops and ignore the refraction introduced by glass. A raindrop is modeled by spherical cap where the radius of the sphere is $ r $ and the angle between tangent and glass plane is $ \tau $. These two parameters determine the volume of the raindrop in glass. If a light ray determined by origin and the location of a pixel in image plane does not go through any raindrops, we set the pixel value unchanged as the background pixel. On the contrary, if a light ray goes through a raindrop in glass plane, we track the light ray by considering the refraction introduced by the raindrop, and set the pixel to the crossover point of light ray and the background plane. In Fig.~\ref{fig:refraction}, the light ray represented by the green line does not go through any raindrop, so we keep the corresponding pixel in image plane unchanged. The light ray represented by yellow line is refracted twice and reaches the same point in the background plane as the green line, so we set the corresponding pixel in image plane same as the green line. If total reflection happened when the light ray propagates from the inside of a raindrop to the air, we set the corresponding pixel in image plane to black. This phenomenon is quite common at real world raindrop's boundary which called dark bands \cite{you2016waterdrop}.

\begin{figure}[t]
	\begin{center}
		\includegraphics[width=1.0\linewidth]{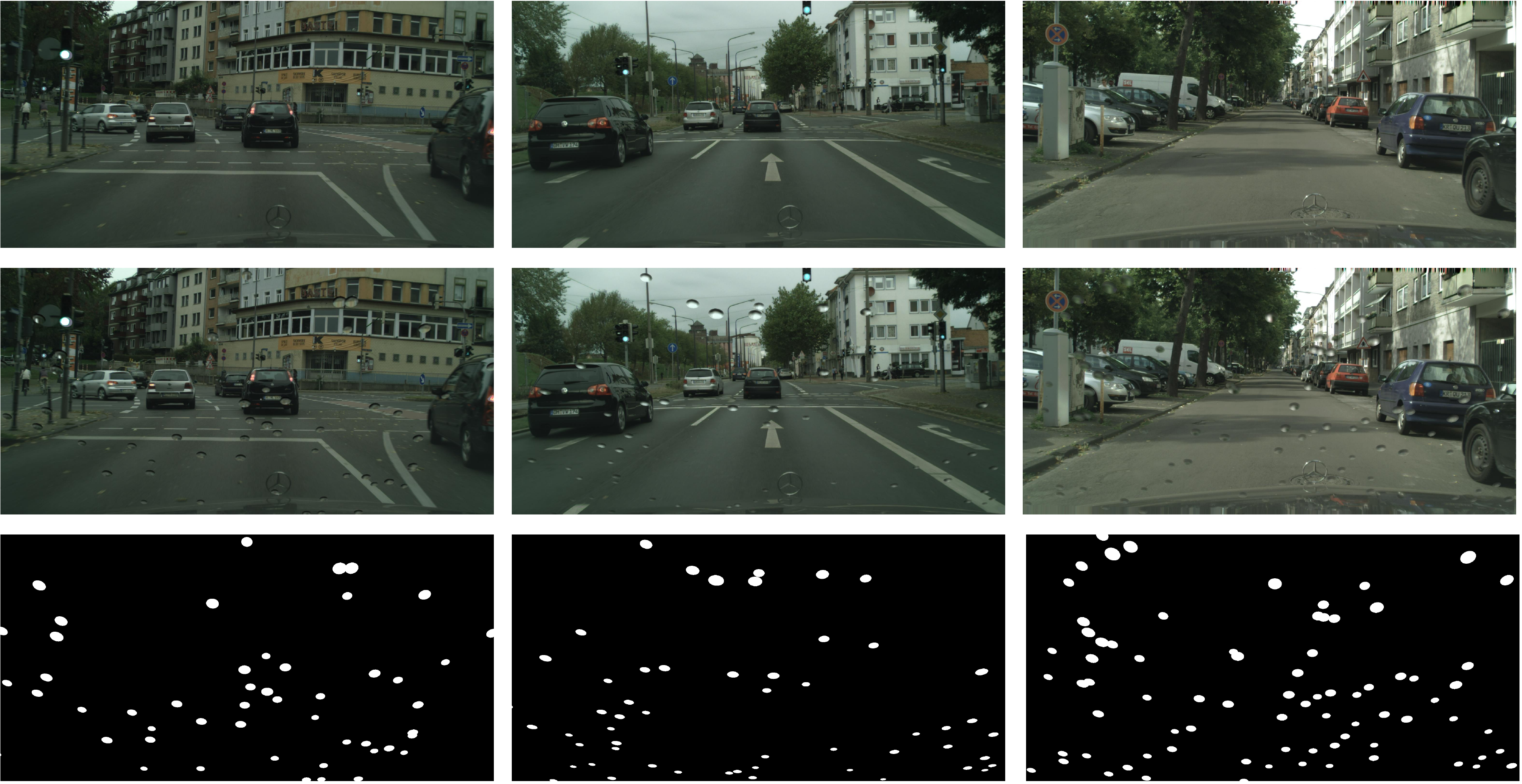}
	\end{center}
	\caption{Samples of our synthetic raindrop images. Top: The ground truth clear image in Cityscapes dataset \cite{cordts2016cityscapes}. Middle: The synthetic raindrop image produced by our refraction model. Bottom: The ground truth binary mask of the raindrops.}
	\label{fig:data_example}
\end{figure}

\paragraph{Dataset Generation: Blurring and Blending.} In the real world, raindrops will be blurred when a camera focuses on the environment scene. We use a disk blur kernel to blur the areas occupied by raindrops in synthetic image. As we observed, it is more realistic for the scene in our dataset to set the diameter of the disk blur kernel to be $ 7\sim 20 $ pixels. Since we already know the locations of raindrops on glass, it is also very convenient to get the ground truth pixel-level binary mask of raindrop image.

\paragraph{Dataset Generation: Environment Realness.} We use images in Cityscapes \cite{cordts2016cityscapes} as the background images. Unlike the dataset created by Qian~\etal \cite{qian2018attentive} which is based on campus scenes, the scenes in Cityscapes are mainly focus on urban street where most outdoor vision systems work. And there are many data recorded in cloudy weather in Cityscapes, while the data in Qian's dataset is recorded in fine weather. So our dataset is more suitable for raindrop removal in outdoor vision systems especially autonomous driving.

\paragraph{Summary of the dataset:} In order to make the raindrop appearance close to real ones, we set $N \in [20, 40]$, $T \in [800, 1500]$, $ r \in [0.8, 1.5]$, $ \psi \in [\ang{30}, \ang{45}] $ and $ \tau \in [\ang{30}, \ang{45}] $. For each background image, we generate $50$ to $70$ raindrops. Finally, we make a dataset containing about $30000$ images based on the training set of Cityscapes for training and $1525$ images based on the test set of Cityscapes for testing.

\section{End-to-End Raindrop Detection and Removal Network}
\label{sec:mehtod}

\begin{figure*}[ht]
	\begin{center}
		\includegraphics[width=\linewidth]{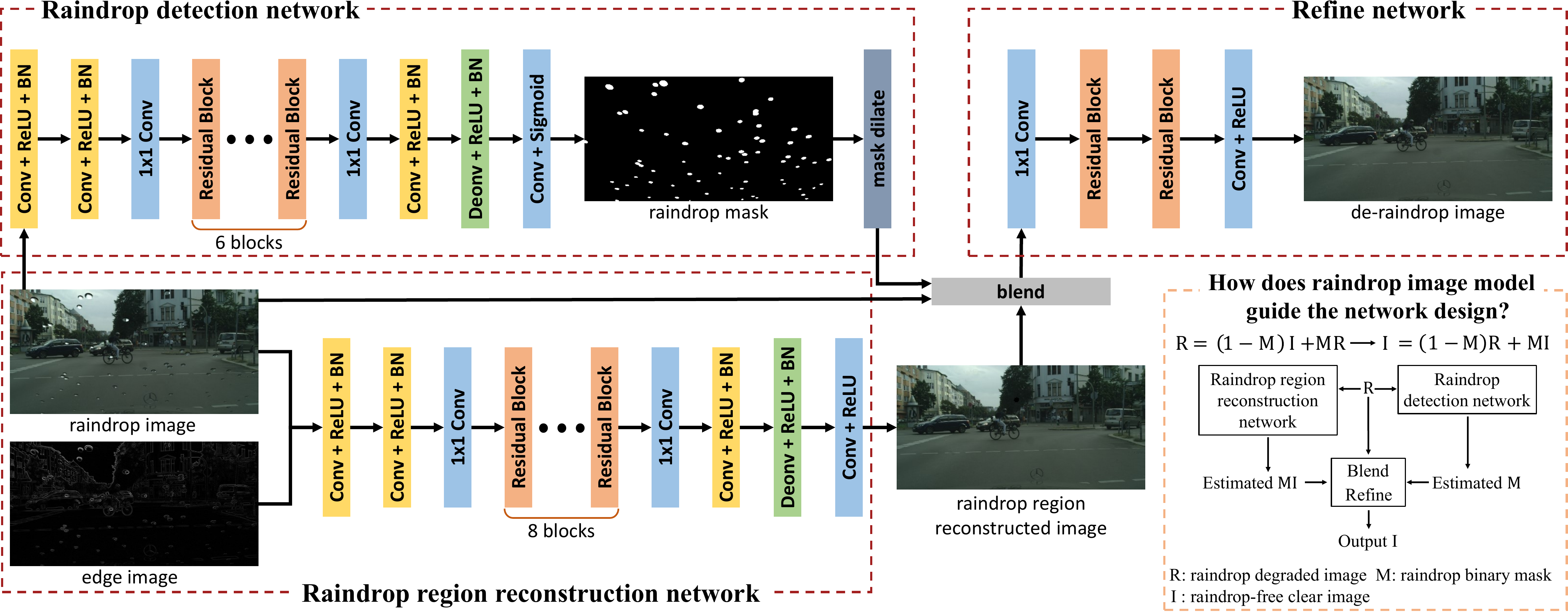}
	\end{center}
	\caption{Network architecture of our proposed method. The whole architecture consists of three sub-networks for raindrop detection, raindrop region reconstruction and refining respectively.}
	\label{fig:network}
\end{figure*}

We devise an end-to-end multi-task network which explicitly incorporate the raindrop imagery model. The observed raindrop degraded image $\mathbf{O}$ can be modeled as $ \mathbf{O} = (1 - \mathbf{M})\mathbf{B} + \mathbf{R} $. Where $ \mathbf{M} $ is raindrop binary mask, $ \mathbf{B} $ is the clear background image and $ \mathbf{R} $ is the raindrop layer. Based on this model, it is intuitive to separate the difficult task into three sub-problems: the first sub-network of our proposed method is designed to detect the raindrop binary mask $ \mathbf{M} $, the second is designed to restore the regions occupied by raindrops, the third is designed to smooth and refine the blended image. As shown in Fig.~\ref{fig:network}, our proposed raindrop removal network consists of three sub-networks to address the three sub-problems respectively. In this section, we first introduce these sub-networks in detail. Then, by combining all sub-networks, we describe the whole architecture of our proposed network and some implementation details.

\subsection{Raindrop Detection Network}

The purpose of our raindrop detection network is to detect the areas of raindrops from input image. The network outputs a pixel-level binary mask in which the pixels of raindrops are marked as ones and the pixels of raindrop-free background are marked as zeros. We can separate the raindrops layer from the background layer by leveraging this binary mask.

Our raindrop detection network is inspired by I-CNN \cite{fan2017generic} and contains stacked residual blocks \cite{he2016deep, he2016identity}. Different from the general semantic segmentation or detection networks \cite{yu2015multi, he2017mask, chen2018deeplab}, the binary mask of raindrops has little semantic information. Hence, we just downsample the internal feature maps to half size in order to enlarge the receptive field. It makes the feature maps denser and keeps more accurate location information.

As shown in Fig.~\ref{fig:network}, the proposed raindrop detection network has $5$ convolution layers and $6$ residual blocks. In the second convolution layer which with stride 2, the resolution of feature maps is reduced to the half of input image. There is a $1 \times 1$ convolution layer in which the channels of feature maps increase from $64$ to $256$. In order to reduce the training time and memory usage, we use residual block in bottleneck fashion. The residual block consists of two $1 \times 1$ and one $3 \times 3$ convolution layers, where the $1 \times 1$ layers will reduce/increase the channels of internal feature maps to $64$/$256$ respectively, and the middle $3 \times 3$ layer has 64-dimensional feature maps in both input and output. All convolution layers in our proposed network are followed by batch normalization (BN) \cite{ioffe2015batch} and ReLU \cite{nair2010rectified}. We use the binary cross-entropy as loss function of the raindrop detection network, and the loss defined as:
\begin{equation}
\medmuskip=-1mu
\thinmuskip=-1mu
\thickmuskip=-1mu
\small
\mathcal{L}_{det}(M, \hat{M}) = -\frac{1}{n} \sum_{i}^{n}\left[M_i\log(\hat{M}_i)+ (1-M_i)\log(1-\hat{M}_i)\right],
\end{equation}
where $M$ is the ground truth binary mask, $\hat{M}$ is probability mask predicted by our network, $n$ is the number of pixels in mask, and $i$ is pixel index.

\subsection{Raindrop Region Reconstruction Network}

The raindrop region reconstruction network is designed to recover the areas occupied by blurred raindrops according to the contextual information, and it shares the similar CNN architecture with the proposed raindrop detection network. Different from the raindrop detection network, we increase the number of residual blocks from $6$ to $8$. We combine the input image and the edge of input image to a $4$-channel tensor as the input. The edge cues can help tasks like reflection removal and image smoothing according to \cite{levin2007user, li2013exploiting, xu2015deep}. We compute the edge image $E$ of a raindrop image $R$ by the equation defined as:
\begin{align}
E_{x, y} &= \frac{1}{4}\sum_{c} (|R_{x, y, c} - R_{x+1, y, c}| + |R_{x, y, c} - R_{x-1, y, c}|\nonumber\\
&+ |R_{x, y, c} - R_{x, y+1, c}| + |R_{x, y, c} - R_{x, y-1, c}|),
\end{align}
where $x, y$ are pixel coordinates, and $c$ is the color channels in RGB image.

The loss function of raindrop region reconstruction is defined as:
\begin{equation}
\mathcal{L}_{recons}(I, \hat{I}) = \frac{1}{n} \sum_{i}^{n} \lambda_i |I_i - \hat{I}_i |,
\end{equation}
where $I$ is the ground truth clear image and $\hat{I}$ is the image predicted by our network. The $\lambda_i$ is a weight which is set to $20$ when pixel $i$ belongs to a raindrop, otherwise to $1$. By introducing $\lambda$, our network will pay more attention to reconstruct the raindrop region.

\subsection{Refine network}

Combing the two sub-networks described above, we propose the refine network. The blended input image $B$ of refine network is defined as:
\begin{equation}
B = \hat{M}\hat{I} + (\mathbf{1} - \hat{M})R,
\end{equation}
where $R$ is raindrop image, $\hat{M}$ is binary mask produced by raindrop detection sub-network, and $\hat{I}$ is the output of raindrop region reconstruction sub-network. $B$ consists of background pixels in $R$ and reconstructed pixels in $\hat{I}$. The architecture of refine network is relatively simple, it contains two convolution layers and two residual blocks. To train the refine network by considering both image structure similarity and color similarity \cite{zhao2017loss}, we use loss function mixed with SSIM \cite{wang2004image} loss and $\ell_1$ loss.
\begin{equation}
\mathcal{L}_{ref}(I, \tilde{I}) = \alpha(1 - \mathcal{L}_{ssim}(I, \tilde{I})) + (1 - \alpha)\mathcal{L}_{\ell_1}(I, \tilde{I}),
\end{equation}
where $\tilde{I}$ is output of our refine network (\ie final output of our proposed method). We set the $\alpha = 0.3$.

\begin{figure}
	\centering
	\subfigure[]{
		\hspace{-0.8mm}\includegraphics[width=0.48\linewidth]{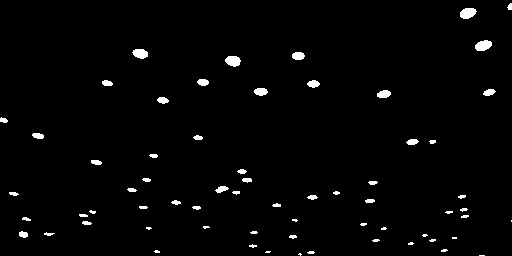}}\hfill
	\subfigure[]{
		\includegraphics[width=0.48\linewidth]{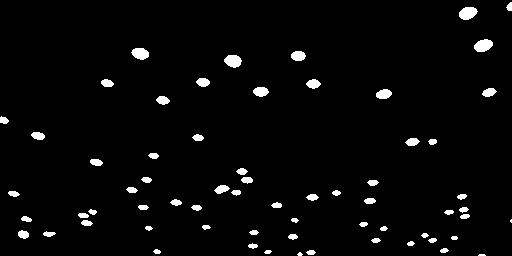}}\vfill\vspace{-2.5mm}
	\subfigure[]{
		\hspace{-0.8mm}\includegraphics[width=0.48\linewidth]{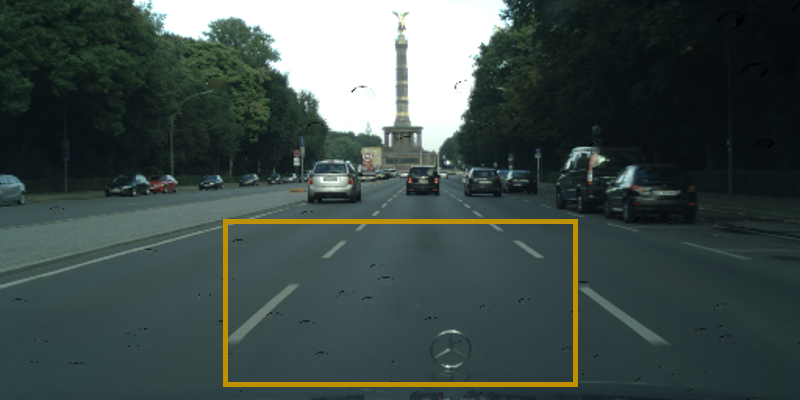}}\hfill
	\subfigure[]{
		\includegraphics[width=0.48\linewidth]{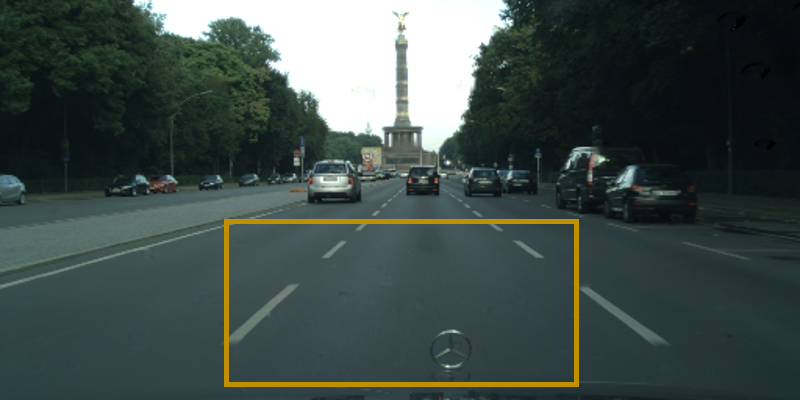}}\vfill\vspace{-2.5mm}
	\subfigure[]{
		\hspace{-0.8mm}\includegraphics[width=0.48\linewidth]{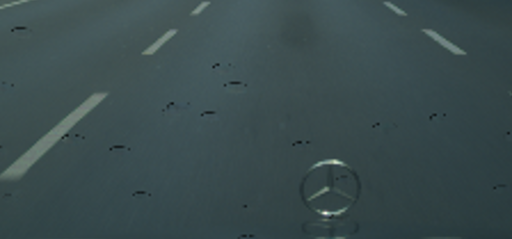}}\hfill
	\subfigure[]{
		\includegraphics[width=0.48\linewidth]{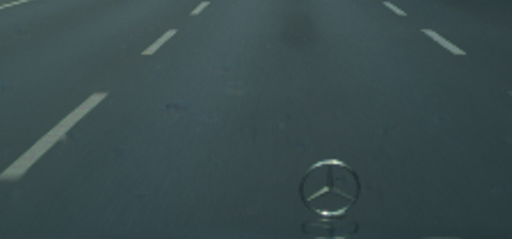}}
	\caption{The effect of dilation operation. \textbf{(a):} the original binary mask produced by raindrop detection network. \textbf{(b):} the dilated binary mask. \textbf{(c):} blended image produced by (a). Note that there are many raindrop edge pixels in the road of the blended image. \textbf{(d):} blended image produced by (b). The raindrop edge pixels in road are removed. \textbf{(e), (f)}: local regions of blended images.}
	\label{fig:blend}
\end{figure}

\paragraph{Dilated Mask:} In experiments, we find that our proposed raindrop detection network gets a relatively low recall at the edge of raindrops. Hence, the $B$ will preserve some raindrop edge pixels if using the original binary mask $\hat{M}$. In order to reduce the number of raindrop pixels in $B$, we apply dilation, a basic mathematical morphology operation, to $\hat{M}$. It is implemented as a parameter-free layer after the output of raindrop detection network. Fig.~\ref{fig:blend} shows the effect of dilation operation on a binary mask and the blended input of refine network.

\subsection{Implementation Details}

\paragraph{Two-Stage Training:} Our complete network consists of three different sub-networks, and we train them in a two-stage fashion. In the first stage, we train raindrop detection network and raindrop region reconstruction network respectively because their outputs are dependencies of the refine network. In the second stage, we combine the networks trained on the first stage with refine network to construct our whole network, and we only update the parameters in refine network when training. Limiting the number of trainable  parameters in the second stage can be beneficial to prevent our network from overfitting.

\paragraph{Data Augmentation:} We do online data augmentation for all training stages. All the images and masks in our training set have resolution of $256 \times 512$. Before feeding into the network, a training pair or its horizontal flip will be randomly cropped to $100 \times 100$.

Our implementation is built on TensorFlow \cite{abadi2016tensorflow} and trained on a single NVIDIA TITAN Xp GPU with 12GB memory. All trainable parameters in proposed networks are initialized by Xavier initializer \cite{glorot2010understanding}. The batch size is set to $32$, and all three sub-networks are trained for $50$ epochs which consists of $50$K steps per epoch. We adopt Adam \cite{kingma2014adam} optimizer with initial learning rate $= 0.001$ which is decayed linearly from $20$ to $40$ epoch until reaching the ending learning rate $= 0.0001$.

\section{Experiments}

In this section, we compare our proposed method to Eigen \cite{eigen2013restoring}, Qian \cite{qian2018attentive}, DID-MDN \cite{zhang2018density} and Pix2Pix \cite{isola2017image} along with ablation experiments. Note that Eigen \cite{eigen2013restoring} and Qian \cite{qian2018attentive} are the only methods in the literature dedicated to this problem  to the best of our knowledge. For this reason, we add a general image-to-image translator Pix2Pix \cite{isola2017image} and a rain streak removal method DID-MDN \cite{zhang2018density} in the comparison. We report the PSNR and SSIM metrics in our synthetic dataset and Qian's dataset.

\subsection{Experiments on Synthetic Image}

\paragraph{Quantitative Evaluation:} Our synthetic test set contains $1525$ rainy/clear images pairs and corresponding rain masks. The test set is synthesized based on the official test set of Cityscapes. We train all the existing methods on our synthetic training set to compare to our proposed method.

\begin{figure}[t]
	\begin{center}
		\subfigure{
			\includegraphics[width=0.3\linewidth]{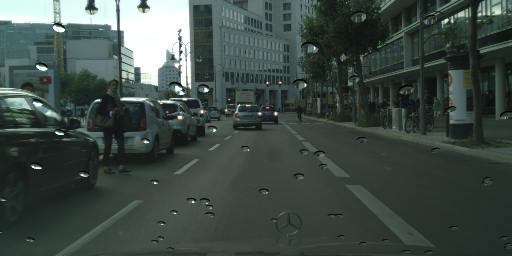}}
		\subfigure{
			\includegraphics[width=0.3\linewidth]{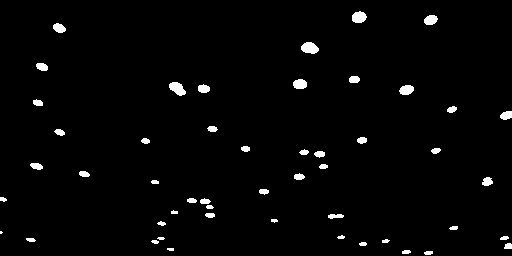}}
		\subfigure{
			\includegraphics[width=0.3\linewidth]{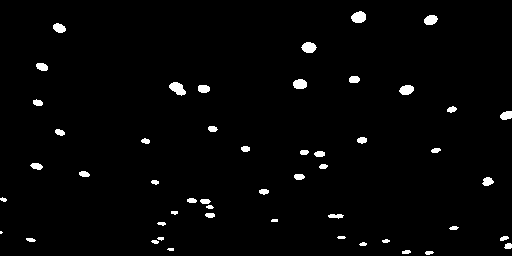}}\vfill\vspace{-2.0mm}
		\setcounter{subfigure}{0}
		\subfigure[Input image]{
			\includegraphics[width=0.3\linewidth]{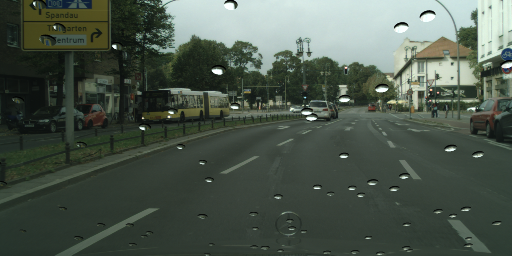}}
		\subfigure[Ground truth mask]{
			\includegraphics[width=0.3\linewidth]{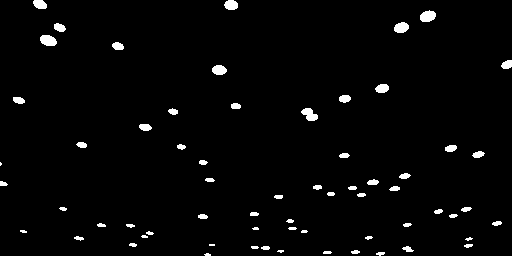}}
		\subfigure[our mask]{
			\includegraphics[width=0.3\linewidth]{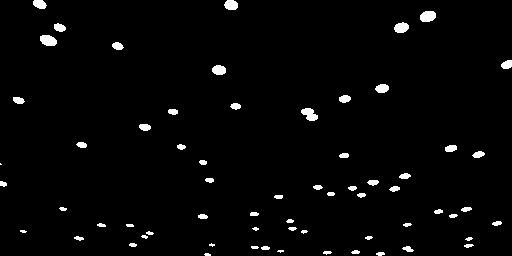}}\vspace{1.0mm}
		\includegraphics[width=1.0\linewidth]{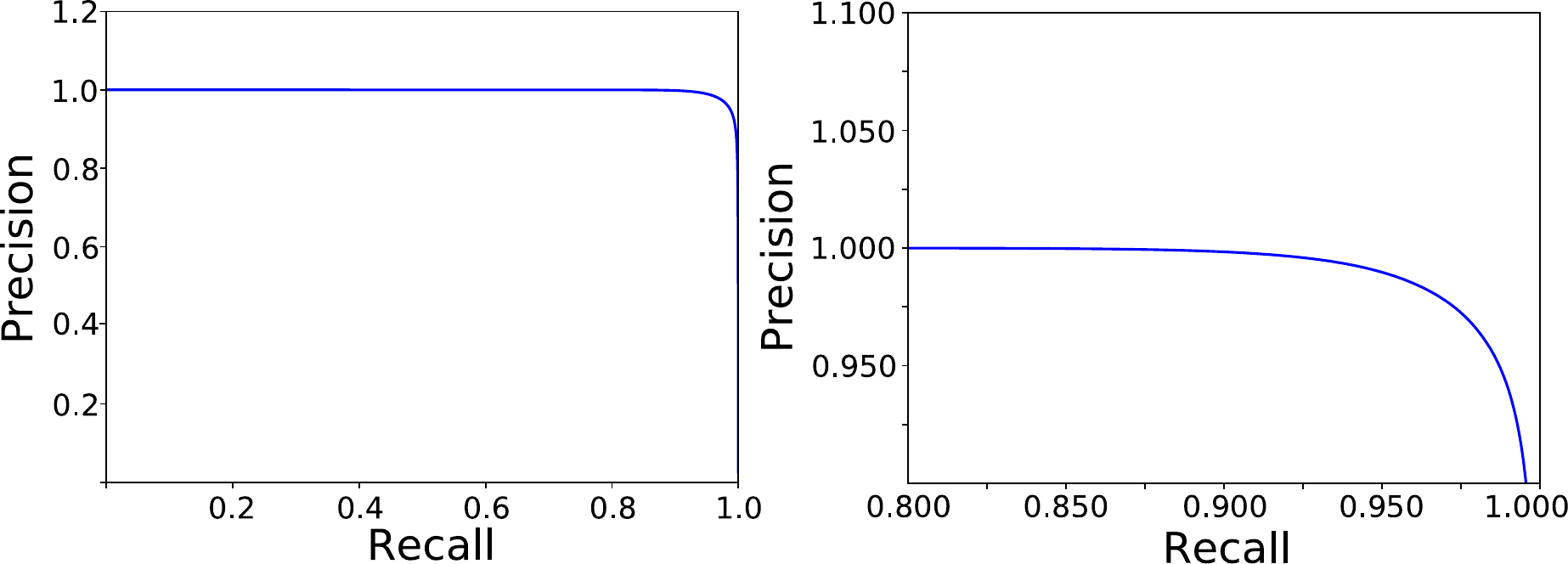}
	\end{center}
	\caption{Raindrop detection on our synthetic dataset. Above: Visual result. Below: Precision-Recall curve. We compute pixel-level accuracy. Curve figure in the right is magnification of figure in the left. Other methods do not output raindrop detection results and are therefore not compared.}
	\label{fig:pr}
\end{figure}

\begin{table}[t]
	\centering
	\begin{tabular}{l|cc}
		& PSNR         & SSIM         \\ \shline
		Eigen \cite{eigen2013restoring} & 29.02 & 0.9560 \\ \hline
		Pix2Pix \cite{isola2017image} & 31.00  & 0.9471       \\ \hline
		DID-MDN \cite{zhang2018density} & 31.32 & 0.9576 \\ \hline
		Qian \cite{qian2018attentive} & 37.79 & 0.9692 \\ \hline
		Ours (3RN only)                    & 37.73        & 0.9852       \\ \hline
		Ours (3RN + RDN + RFN)       & 39.24        & 0.9848       \\ \hline
		Ours (3RN + Dilated RDN + RFN) & \textbf{41.29}     & \textbf{0.9921}       \\
	\end{tabular}
	\vspace{1em}
	\caption{Quantitative evaluation results of raindrop removal on synthetic dataset. \textbf{3RN:} Our Raindrop Region Reconstruction Network. \textbf{RDN:} Our Raindrop Detection Network. \textbf{RFN:} Our Refine Network. The setting in the last row is our complete model.}
	\label{table:syn_compare}
\end{table}

\begin{table}[t]
	\centering
	\setlength{\tabcolsep}{10pt}
	\begin{tabular}{l|cc}
		& PSNR         & SSIM         \\ \shline
		Eigen \cite{eigen2013restoring} & 28.59 & 0.6726 \\ \hline
		Pix2Pix \cite{isola2017image} & 30.14  & 0.8299       \\ \hline
		DID-MDN \cite{zhang2018density} & 27.06 & 0.8830 \\ \hline
		Qian \cite{qian2018attentive}          & \textbf{30.82}        & 0.9050     \\ \hline
		Ours (3RN only)               & 29.28        & 0.9016       \\ \hline
		Ours (rough mask)               & 30.17        & \textbf{0.9128}       \\
	\end{tabular}
	\vspace{1em}
	\caption{Quantitative evaluation results of raindrop removal on dataset proposed in Qian \cite{qian2018attentive}. \textbf{Ours (3RN only):} Our Raindrop Region Reconstruction Network only. \textbf{Ours (rough mask):} Our model trained with \textit{low-quality} rough mask which is produced by subtracting raindrop image with the ground truth.}
	\label{table:real_compare}
\end{table}

\begin{figure*}[t]
	\centering
	\includegraphics[width=1.0\linewidth]{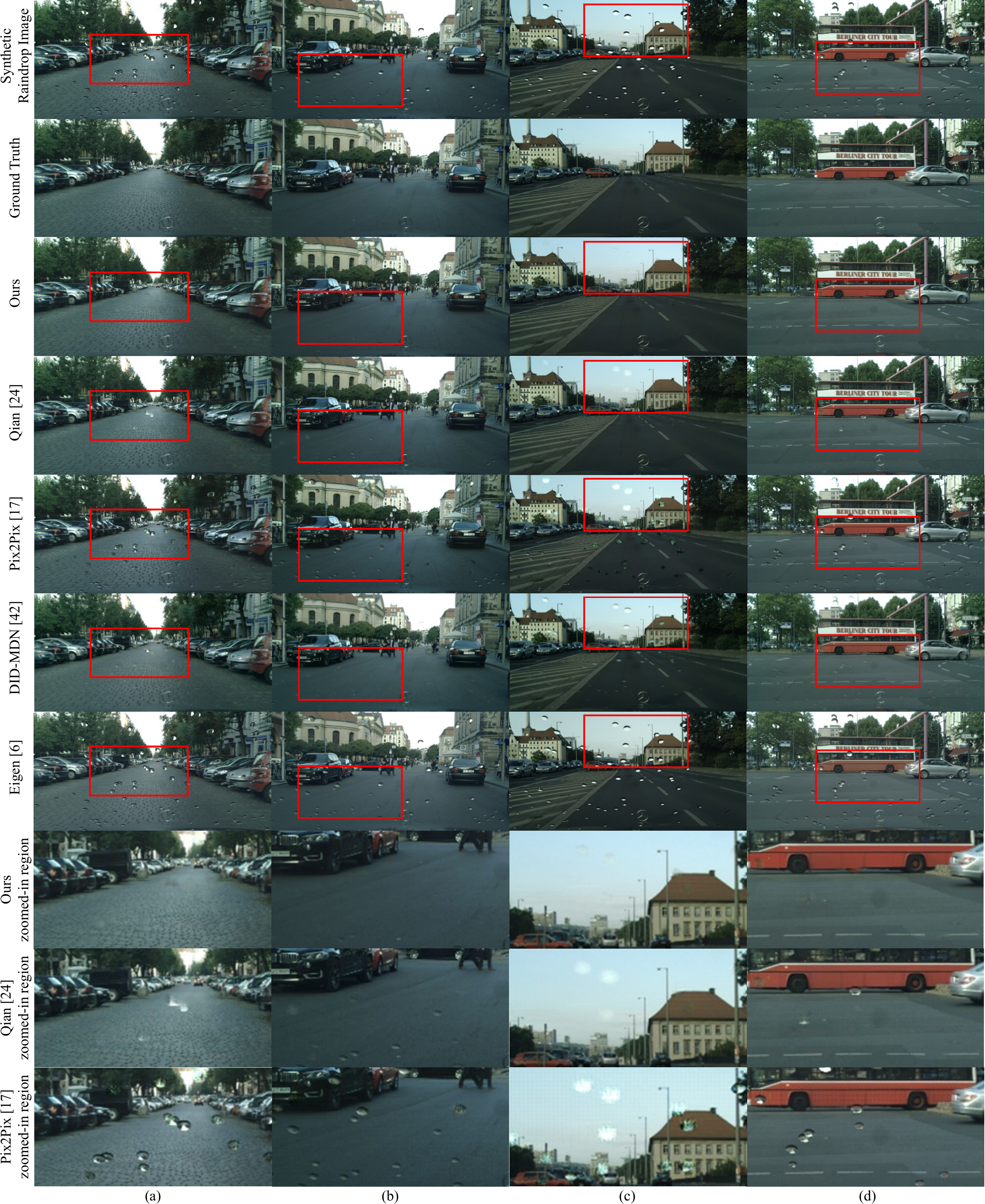}
	\caption{Comparison of raindrop removal on our synthetic dataset. It can be seen that the proposed method achieves better performance in both removing raindrops and avoiding artifacts. Best viewed with zoom.}
	\label{fig:syn_vis}
\end{figure*}
\begin{figure*}[ht]
	\centering
	\includegraphics[width=1.0\linewidth]{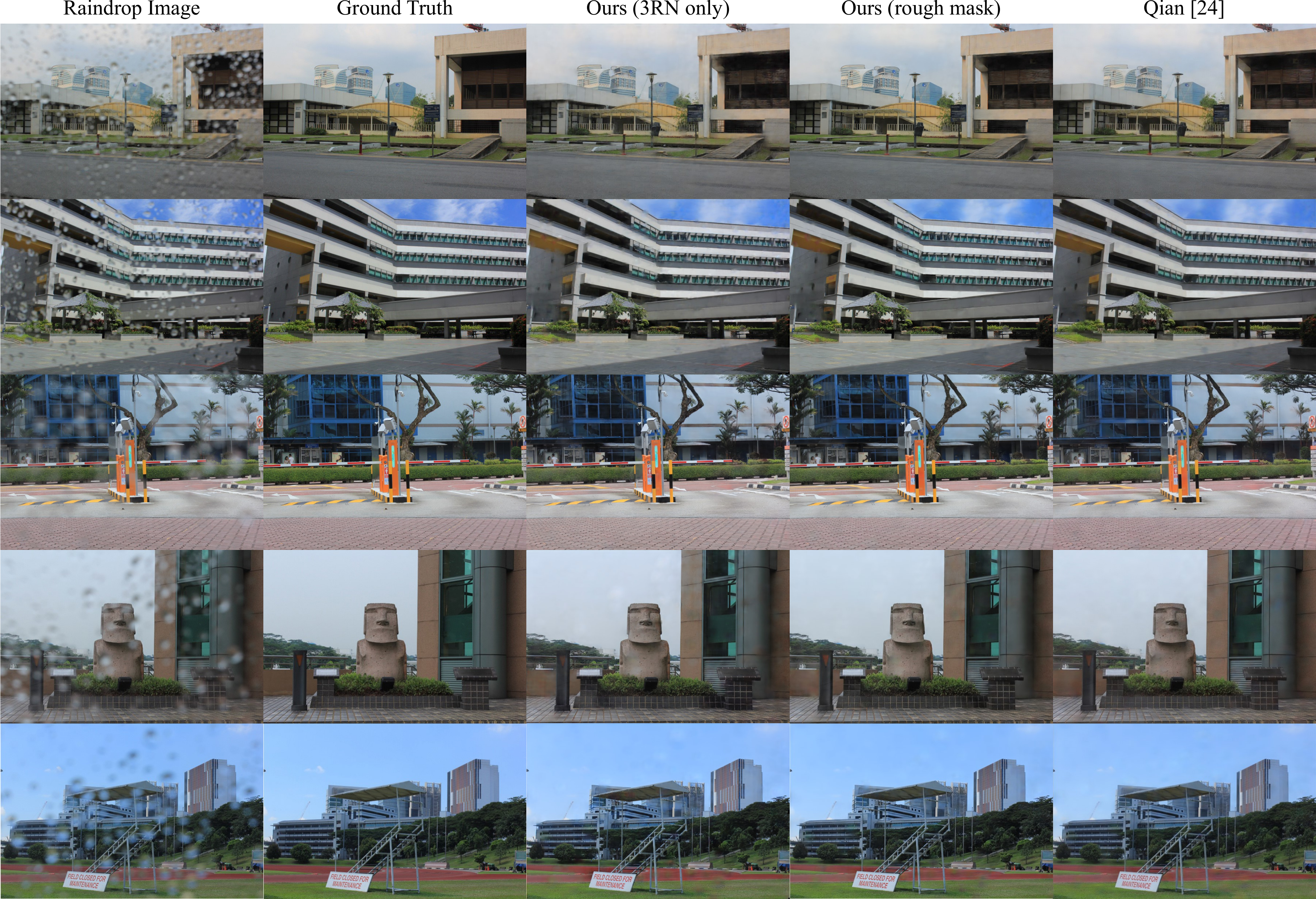}
	\caption{Comparison of raindrop removal results on Qian's dataset \cite{qian2018attentive}. All three models remove raindrops successfully and produce clear images.}
	\label{fig:qian_vis}
\end{figure*}

First, we evaluate the performance of our raindrop detection network on synthetic dataset which contains the ground truth of binary raindrop masks. Fig.~\ref{fig:pr} shows Precision-Recall curve of our raindrop detection network. The average precision (AP) of our network prediction is $0.9973$. It indicates that our raindrop detection network works very well on synthetic images. Because other methods do not predict the raindrop mask, we cannot compare our raindrop detection network to them.

Table~\ref{table:syn_compare} shows quantitative raindrop removal results. It is clearly that our complete model outperforms other methods in terms of both PSNR and SSIM by a large margin.

\paragraph{Qualitative Evaluation:} Fig.~\ref{fig:syn_vis} shows the qualitative results of different methods. Our proposed method, Qian \cite{qian2018attentive} and DID-MDN \cite{zhang2018density} remove the most of raindrops successfully while the results produced by Pix2Pix \cite{isola2017image} and Eigen \cite{eigen2013restoring} still contain many raindrops. In the first row of Fig.~\ref{fig:syn_vis}, there are artifacts in the center of Qian and DID-MDN results. In the second row, there are some raindrops in the bottom of results of Qian and DID-MDN. Our method achieves better performance in both removing raindrops and avoiding artifacts.

\paragraph{Ablation Study:} We run a number of ablations in order to demonstrate the effectiveness of different modules in our proposed method. The quantitative results are shown in Table \ref{table:syn_compare}. We use the raindrop region reconstruction network as baseline. By adding the raindrop detection network and refine network to the baseline, both PSNR and SSIM are increasing. Then, by replacing the mask with dilated mask, we get our complete proposed architecture which archives the highest PSNR and SSIM. The results of ablation study indicate that all our proposed modules do contribute to the raindrop removal. Our complete architecture improve PSNR by 9.4\% compare to the 3RN only. It is a substantially large improvement.

\paragraph{Efficiency:} The running speed is critical in many outdoor computer vision systems. In order to show the efficiency of our method, we also evaluate the inference speed of prior methods and ours. For a single $ 256 \times 512 $ image, the inference time of our method is $ 69.74 $ms, Pix2Pix \cite{isola2017image} is $ 79.1 $ms, Qian \cite{qian2018attentive} is $ 97.16 $ms, Eigen \cite{eigen2013restoring} is 107.712ms and DID-MDN \cite{zhang2018density} is 133.64ms. The results indicate that our method is not only effective but also efficient. All the methods are tested on a single NVIDIA TITAN Xp GPU, and the time reported is the average of $ 20 $ repeats.

\subsection{Experiments on Real-world Image}

We evaluate the proposed method on real-world dataset introduced in Qian \cite{qian2018attentive}. Due to the lack of the ground-truth raindrop mask, we cannot train our complete architecture directly. Instead, we only use \textit{low-accuracy} masks roughly estimated by subtracting raindrop images with ground truth. We also train our Raindrop Region Reconstruction network (3RN) which do not need raindrop masks. As shown in Table \ref{table:real_compare}, our methods get competitive results quantitatively that is on par with Qian \cite{qian2018attentive}. Note the dataset is not very challenging because the data in training set and test set is too similar. Thus all three models can produce compelling visual quality raindrop removal as shown in Fig.~\ref{fig:qian_vis}.

\section{Conclusions}

In this paper, we propose a novel end-to-end network to detect and remove adherent raindrop jointly. Since the supervised deep learning in raindrop removal suffers from the lacking of sufficient paired training data, we also develop a practical and realistic dataset for adherent raindrop which contains the pixel-level raindrop binary masks. There are two stages in our proposed multi-task network. In the first stage, two sub-networks detect raindrop locations and restore adherent raindrop regions respectively. In the second stage, a blended image is produced by using the location clues, then a refine network is employed to smooth the blended image. Our experiment results show that the proposed method outperforms the state-of-the-art and can handle both synthetic data and real-world data. In the future, we would like to enhance our raindrop imagery model and extend our experiments to further validate the benefit of using our method on computer vision systems working under the rainy scenes.

{\small
\bibliographystyle{ieee}
\bibliography{egbib}
}

\end{document}